\documentclass{article}

\usepackage{microtype}
\usepackage{graphicx}
\usepackage{subcaption}
\usepackage{booktabs} 
\usepackage{multirow}
\usepackage{multicol}
\usepackage{hyperref}

\usepackage[preprint]{icml2026}

\usepackage{amsmath}
\usepackage{amssymb}
\usepackage{mathtools}
\usepackage{amsthm}

\usepackage[capitalize,noabbrev]{cleveref}

\theoremstyle{plain}

\theoremstyle{definition}

\theoremstyle{remark}

\usepackage[textsize=tiny]{todonotes}

\icmltitlerunning{Submission and Formatting Instructions for ICML 2026}

\begin{document}

\twocolumn[
  \icmltitle{Efficient Post-training of LLMs for Code Generation With Offline Reinforcement Learning}



  \icmlsetsymbol{equal}{*}

  \begin{icmlauthorlist}
    \icmlauthor{Mingze Wu}{equal,yyy}
    \icmlauthor{Abhinav Anand}{equal,yyy}
    \icmlauthor{Shweta Verma}{yyy}
    \icmlauthor{Mira Mezini}{yyy,sch,comp}
  \end{icmlauthorlist}

  \icmlaffiliation{sch}{Hessian Center for Artificial Intelligence, Darmstadt, Germany}
  \icmlaffiliation{yyy}{Technische Universitat, Darmstadt, Germany}
  \icmlaffiliation{comp}{National Research Center for Applied Cybersecurity ATHENE}

  \icmlcorrespondingauthor{Abhinav Anand}{abhinav.anand@tu-darmstadt.de}

  \icmlkeywords{Machine Learning, ICML}

  \vskip 0.3in
]



\printAffiliationsAndNotice{}  

\begin{abstract}
Post-training using online reinforcement learning (RL) is an important training step for LLMs, including code-generating models. However, online RL for code generation involves LLM inference and verification of the generated output, which can take considerable time and resources. In this paper, we explore the application of offline RL to code-generating models by leveraging existing code datasets. Our experiments demonstrate that offline RL is an effective training strategy for improving LLM performance. We show that offline RL can be especially beneficial for small LLMs and challenging coding problems.

\end{abstract}

\section{Introduction}
Training large language models (LLMs) with feedback that reflects human preferences is a central component of improving model performance and aligning outputs with user expectations. In domains such as software engineering and mathematics, these preferences are often grounded in verifiable rewards. For example, in software engineering, functional correctness of generated code is the most widely used criterion for validating model outputs.

Numerous reinforcement learning (RL) algorithms --such as PPO \cite{ppo}, GRPO \cite{grpo}, RLOO \cite{rloo, rloollm}, and their variants -- have been proposed to train LLMs from preference feedback. These methods are inherently on-policy and rely on online data collection. In practice, however, training is often bottlenecked by the high cost of transformer inference. To mitigate this, high-throughput inference engines such as SGLang or vLLM \cite{vllm} are commonly used. This introduces a discrepancy between training and inference logits, effectively making the training process off-policy \cite{offpolicy}. Empirically, the observed performance gains in this setting suggest that on-policy algorithms can still be successfully applied in mildly off-policy regimes, at least for LLMs.

In this work, we take this observation further by systematically studying the behavior of online policy gradient methods in a fully offline setting. Specifically, we train models on pre-collected datasets instead of sampling new data from the model. This setup addresses several key limitations of online RL for LLMs:

\textbf{Efficiency.} Offline training removes the need for repeated sampling and external verification (e.g., by executing generated code), substantially reducing training cost and latency.

\textbf{Diversity.} Online RL for LLMs often suffers from entropy collapse \cite{dapo}, where output diversity decreases over time, limiting exploration and diminishing returns. Offline training mitigates this issue by leveraging pre-existing datasets with inherent diversity; additionally, diversity can be explicitly controlled through data selection or batching strategies.

\textbf{Training stability.} In online settings, when all sampled solutions for a prompt are correct, the resulting low reward variance can lead to unstable or uninformative gradient updates, sometimes degrading performance. Offline data, which naturally includes a mix of solution qualities, helps maintain informative training signals.



Our empirical study shows that:

\begin{itemize}
    \item Online policy gradient algorithms can be directly applied in an offline setting to improve code generation performance. 
    \item Offline RL yields the largest gains on challenging programming problems.
    
    \item 
    Offline training improves pass@10—particularly on harder tasks, indicating that it preserves or even enhances output diversity.
    
\end{itemize}

At the same time, we identify important limitations. When the base model already performs well, offline RL provides little to no improvement. Similarly, for easier problems, supervised fine-tuning (SFT) can outperform offline RL. These findings suggest the need for improved algorithms, potentially combining offline RL with value-based methods such as Q-learning, or more adaptive training strategies.

\begin{figure*}[h]
  \vskip 0.2in
  \begin{center}
    \centerline{\includegraphics[width=\linewidth]{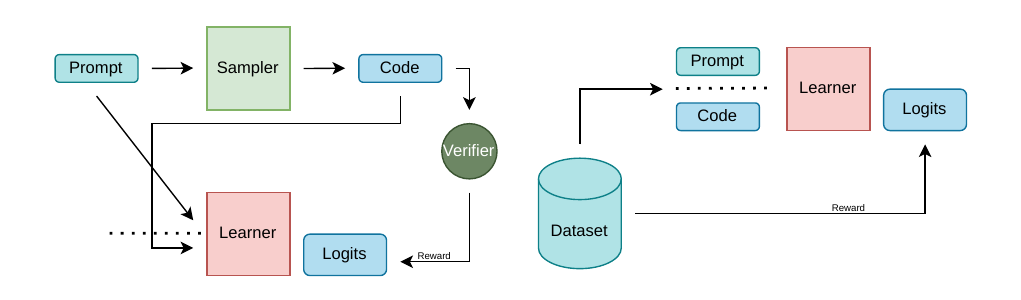}}
    \caption{
      Comparison of typically used setup for training LLM with rewards (left) v/s the proposed offline setup (right). In left, the sampler and learner are the same models but running in different modes for inference and training respectively resulting in deviation between sampling policy and the policy being trained. The proposed setup removes the need for sampling and verification during training, making training faster and efficient.
    }
    \label{offline_desc}
  \end{center}
\end{figure*}

\section{Training Setup}
\subsection{Models, Training Method, and Dataset}
We use Qwen2.5-Coder \cite{qwen} as our base model. We consider two scales, with 0.5B and 7B parameters. For the 0.5B model, we fine-tune all parameters, whereas for the 7B model we apply LoRA \cite{lora} adapters instead of full-model training.

We train on the CodeNet dataset \cite{codenet}. CodeNet is a large-scale corpus comprising around 4k programming problems and 14 million solutions in more than 50 programming languages, accompanied by rich metadata. This metadata includes functional correctness, syntactic correctness, runtime, memory usage, and code size, which can be used to define reward signals for code quality. In this work, we restrict ourselves to Python solutions and use only functional and syntactic correctness as feedback.

A key limitation of CodeNet is its imbalance. This appears along two axes: (1) the number of solutions per problem, which ranges from 0 to 27k, and (2) a very high proportion of correct solutions. To partially mitigate this, we cap the number of correct solutions per problem at 50.

\subsection{Offline RL with Leave one out}
Following \citet{rloollm}, we adopt the REINFORCE Leave-one-out (RLOO) objective \cite{rloo}, defined as: 

\begin{equation}
    \nabla_\theta J_{\text{RLOO}}(\theta)
    = \frac{1}{n} \sum_{i=1}^{n} \hat{A}_i \,\nabla_\theta \log \pi_\theta(y_i \mid x)
    \label{eq:RLOO_grad}
\end{equation}

where, $\pi_\theta$ is the policy being trained and $\hat{A}_i$ denotes the advantage for sample $i$. In \citet{rloollm}, the sequences $y_1, \dots, y_n$ are drawn from the current policy $\pi_\theta$. In our offline setting, we can assume that these sequences have been sampled from an unknown policy $\pi_\beta$. The advantage is calculated by using $n$ samples for each problem statement to reduce variance. The RLOO advantage used in \citet{rloollm} is 

\begin{equation}
    \hat{A}_i = R(y_{(i)}, x) - \frac{1}{n-1} \sum_{j \neq i} R(y_{(j)}, x) \label{eq:RLOO adv}
\end{equation}

that is, the reward of sample  minus the mean reward of the remaining  samples for the same input.

We also experiment with the advantage formulation commonly used in GRPO, given by

\begin{equation}
    \hat{A}_i^{GRPO} = \frac{R(y_{(i)},x) - mean(\mathbf{R})}{std(\mathbf{R})}
\end{equation}

where $\mathbf{R}$ denotes the vector of rewards for all  samples of a given problem.

\citet{awac} propose an online RL algorithm that combines offline data augmented with online interactions. 
%
%
Although we do not perform online interaction, our method is inspired by their weighting scheme. However, instead of learning a separate critic for advantage estimation, we adopt the GRPO-style advantage:

\begin{equation}
    \hat{A}_i^{exp} = exp(\hat{A}_i^{GRPO})
\end{equation}


This weighting increases the relative contribution of high-reward (correct) samples and down-weights low-reward (incorrect) ones. In effect, the objective moves closer to supervised fine-tuning (SFT) while still leveraging negative examples through their reduced, but non-zero, weights.

\subsection{Batches and Groups}
We experiment with group sizes of 4 and 8 and observe that the models performed better with group size of 4; all subsequent experiments therefore use a group size of 4. 
For each group, we enforce the presence of at least one correct and one incorrect solution. This constraint excludes problems with no correct solutions -- typically the hardest instances -- making it somewhat restrictive, but it substantially reduces the variance of the advantage estimates. In future work, this requirement could be relaxed by incorporating additional variance-reduction techniques, such as advantage clipping.



\subsection{Reward}
We provide the following reward based on the status of the code.

\begin{equation}\label{eq:reward_function}
r =
\begin{cases}
+1.0 & \text{All Test Cases Passed} \\
-0.1 & \text{Test Cases Failed} \\
-0.5 & \text{Time Limit Exceeded} \\
-0.6 & \text{Runtime Error} \\
-1.0 & \text{Compile Error} 
\end{cases}
\end{equation}

\subsection{Computational Constraints}
We frame our study under explicit computational constraints. 
Recent RL algorithms for LLMs typically assume substantial compute budgets, whereas we restrict each training run to a single A100 80GB GPU and limit wall-clock time to under 30 hours for the 7B model and under 5 hours for the 0.5B model.



To meet these constraints, we use a small batch size of 8 and train for a relatively modest number of optimization steps. We further cap the generated code length at 2048 tokens, discarding samples that exceed this limit.



\subsection{Benchmarks}
We evaluate our models on two benchmarks: 1) MBPP \cite{mbpp} and 2) APPS+ \cite{apps}. MBPP consists of simple Python problems, whereas APPS+ includes problems spanning three different difficulty levels:  ``introductory", ``interview", and ``competitive". For MBPP, we report pass@1; for APPS+, we report both pass@1 and pass@10 over all three difficulty categories.

\begin{table*}[ht]
  \caption{Performance of Qwen-2.5-Coder 0.5B model under different training on the APPS+ dataset. The best performance in a category is \textbf{bold}. The performance on competitive problems is trivial across all training setup.}
  \label{table apps 500}
  \begin{center}
    \begin{small}
      \begin{sc}
        \begin{tabular}{lccc|ccc}
          \toprule
          \multirow{2}{*}{Model}  & \multicolumn{3}{c|}{Pass@1} & \multicolumn{3}{c}{Pass@10}   \\
          & Introductory & Interview & Competition & Introductory  & Interview & Competition \\
          \midrule
          Base &0.03&0.06&0.00&0.30&0.60&0.00 \\
          SFT &\textbf{1.94}&1.87&\textbf{0.16}&\textbf{5.9}&8.1&\textbf{0.87} \\
          RLOO &0.06&0.12&0.05&0.5&1.1&0.35 \\
          with GRPO advantage &0.07&0.59&0.05&0.4&3.5&0.35 \\ 
          advatnage with exp() &1.67&\textbf{2.15}&0.10&5.4&\textbf{8.6}&0.70 \\
          \bottomrule
        \end{tabular}
      \end{sc}
    \end{small}
  \end{center}
  \vskip -0.1in
\end{table*}

\begin{table*}[ht]
  \caption{Performance of Qwen-2.5-Coder 7B model under different training on the APPS+ dataset. The best performance in a category is \textbf{bold}. If the performace degrades compared to base model, then it is marked in \color{red}{red}.}
  \label{table apps 7b}
  \begin{center}
    \begin{small}
      \begin{sc}
        \begin{tabular}{lccc|ccc}
          \toprule
          \multirow{2}{*}{Model}  & \multicolumn{3}{c|}{Pass@1} & \multicolumn{3}{c}{Pass@10}   \\
          & Introductory & Interview & Competition & Introductory  & Interview & Competition \\
          \midrule
          Base &4.5&7.01&1.42&11.90&25.70&8.22 \\
          SFT &\textbf{9.21}&11.30&\color{red}{1.38}&\textbf{16.50}&\textbf{30.30}&\color{red}{6.29} \\
          RLOO &6.98&10.31&2.06&13.80&27.80&\color{red}{8.04} \\
          with GRPO advantage &7.72&\textbf{12.14}&\textbf{2.67}&14.70&30.20&\textbf{11.54} \\ 
          advatnage with exp() &7.14&11.09&2.38&15.10&\textbf{30.30}&9.97 \\
          \bottomrule
        \end{tabular}
      \end{sc}
    \end{small}
  \end{center}
  \vskip -0.1in
\end{table*}

\section{Results and Discussion}
In this section, we present the results in \cref{table apps 500}, \cref{table apps 7b}, and \cref{table-mbpp} and discuss them in detail.

\subsection{Offline RL improves code generation capability.}
Across both benchmarks and different difficulty levels of APPS+, we observe that offline RL consistently improves the model’s code generation performance.
The gains are most robust when RLOO is combined with the GRPO-style advantage, underscoring the importance of variance reduction in advantage estimation.

However, we see no improvement for the 7B model on MBPP, suggesting that offline RL offers limited benefits when the base model already performs strongly on a task. 
Enhancing the offline dataset and incorporating a small amount of online training may be necessary to unlock further gains in such regimes.


\subsection{Offline RL improves model on difficult problems.}

A key benefit of our offline training setup is improved performance on difficult problems. We observe gains on interview-level tasks for the 0.5B model and on both interview- and competition-level tasks for the 7B model. The effect is more pronounced on competition-level problems with the 7B model: supervised fine-tuning degrades performance on both pass@1 and pass@10, whereas RLOO with GRPO-style advantages improves pass@1 by 88\% and pass@10 by 40\%.

\subsection{Small model requires different scale of reward.}

As shown in \cref{table apps 500}, RLOO with exponential advantage leads to significant improvements in the model's performance on introductory and interview-level problems. However, unlike the 7B model, neither RLOO nor RLOO with GRPO advantage improves the model. Using the exponential advantage changes the reward range from $[-1, 1]$ to $[0.37, 2.72]$. Thus, small models might benefit from a different reward level. However, the same reward harms the model on MBPP. The impact of reward across various models and benchmarks also underscores the importance of empirically determining the optimal reward during training.

\begin{table}[ht]
  \caption{Performance of 0.5B and 7B model on MBPP benchmark. The best performance is marked in \textbf{bold}. If the performance degrades over the base model, then it is marked in \color{red}{red}.}
  \label{table-mbpp}
  \begin{center}
    \begin{small}
      \begin{sc}
        \begin{tabular}{lcc}
          \toprule
          Training  & 0.5B         & 7B  \\
          \midrule
          Base&36.2&64.4\\
          SFT&\color{red}{35}&\color{red}{63.2}\\
          RLOO&37.4&64.4\\
          with GRPO Advantage&\textbf{39.4}&\color{red}{64}\\
          Advantage with exp()&\color{red}{35}&64.40\\
          \bottomrule
        \end{tabular}
      \end{sc}
    \end{small}
  \end{center}
  \vskip -0.1in
\end{table}

\section{Related Works}
Code generation is a common application area for LLMs, and many open-source LLMs have been released for this task \cite{qwen, deepseekcoder, codegensurvey}. Within this domain, Reinforcement Learning with Verifiable Reward (RLVR) \cite{rlvr} has emerged as an important paradigm, where feedback is derived from verifiable signals (such as test outcomes) rather than from a learned reward model.
However, training is typically carried out in an online fashion.

CodeRL \cite{coderl} is an earlier approach that also leverages verifiable rewards but relies on offline data.  In CodeRL, candidate solutions are first generated by the model and then used for training, which can be problematic when the base model is weak, as it may fail to produce sufficiently good samples. 
Moreover, CodeRL employs a frozen critic network for value estimation. In contrast, we train directly on human-written solutions, which ensures the presence of both good and bad examples and yields higher code diversity, and our method does not require a separate critic model.



\section{Limitations and Future Work}
In this paper, we have shown that offline RL is a promising approach for training LLMs with verifiable rewards for functionally correct code generation. Nonetheless, our study has several limitations:

\begin{itemize} 
    \item We did not systematically explore the space of learning rates, reward formulations, or batch sizes, and in particular relied on relatively small batches.
    \item The underlying dataset is highly imbalanced; although we applied simple heuristics to mitigate this, more principled data filtering and reweighting strategies could further reduce imbalance.
    \item Our experiments are purely offline. Combining offline training with a limited amount of online interaction may improve exploration and overall performance.
    \item We directly apply on-policy online algorithms in an offline setting. 
    Designing theoretically grounded offline RL methods tailored to LLMs, or leveraging value-based approaches such as Q-learning \cite{ilql}, could yield further gains.
    
\end{itemize}

The above points highlight several promising directions for future work along optimization, data curation, training regimes, and algorithmic design.

\section{Conclusion}
We have shown that offline RL is an effective and efficient post-training strategy for code-generating LLMs. It improves both pass@1 and pass@10, with pronounced gains on the most challenging problems. Notably, these improvements are achieved under strict computational constraints and with a highly imbalanced dataset.

There remains substantial room for improvement through better reward design and more carefully curated training data. Incorporating additional signals, such as runtime efficiency, memory usage, and security properties, alongside functional correctness could further enhance code quality. Such multi-faceted evaluation is often easier to implement in an offline setting, where candidate solutions can be analyzed along many dimensions without incurring the cost of repeated online execution.







\bibliography{example_paper}
\bibliographystyle{icml2026}

\newpage

\end{document}